\documentclass[10pt,twocolumn,letterpaper]{article}

\usepackage{iccv}
\usepackage{times}
\usepackage{epsfig}
\usepackage{graphicx}
\usepackage{amsmath}
\usepackage{amssymb}

\usepackage{mathtools}
\usepackage{dsfont}
\usepackage[pagebackref=true,breaklinks=true,letterpaper=true,colorlinks,bookmarks=false]{hyperref}

\usepackage[accsupp]{axessibility}  

\usepackage{color, colortbl}
\usepackage[table]{xcolor}

\renewcommand*{\ie}{i.e.\@\xspace}
\renewcommand*{\eg}{e.g.\@\xspace}

 \iccvfinalcopy 

\def\iccvPaperID{8743} 

\ificcvfinal\fi

\begin{document}

\title{Focus on the Positives: Self-Supervised Learning for Biodiversity Monitoring}

\author{Omiros Pantazis$^{1}$\hspace{15pt}Gabriel J. Brostow$^{1,2}$\hspace{15pt}Kate E. Jones$^{1}$\hspace{15pt}Oisin Mac Aodha$^{3}$\\
$^{1}$University College London \hspace{15pt} $^{2}$Niantic \hspace{15pt} $^{3}$University of Edinburgh\\\url{www.github.com/omipan/camera_traps_self_supervised} 
}

\maketitle
\ificcvfinal\thispagestyle{empty}\fi

\begin{abstract}
We address the problem of learning self-supervised representations from unlabeled image collections. 
Unlike existing approaches that attempt to learn useful features by maximizing similarity between augmented versions of each input image or by speculatively picking negative samples, we instead also make use of the natural variation that occurs in image collections that are captured using static monitoring cameras. 
To achieve this, we exploit readily available context data that encodes information such as the spatial and temporal relationships between the input images. 
We are able to learn representations that are surprisingly effective for downstream supervised classification, by first  identifying high probability positive pairs at training time, \ie those images that are likely to depict the same visual concept. 
For the critical task of global biodiversity monitoring, this results in image features that can be adapted to challenging visual species classification tasks with limited human supervision. 
We present results on four different camera trap image collections, across three different families of self-supervised learning methods, and show that careful image selection at training time results in superior performance compared to existing baselines such as conventional self-supervised training and transfer learning.  
\end{abstract}

\section{Introduction}

Learning transferable representations of visual data without requiring explicit semantic supervision at training time is an important and open problem in computer vision.  
Recent progress on this front has been impressive, resulting in self-supervised methods that are capable of learning features that approach, and in some cases even surpass, their fully supervised counterparts across a range of downstream tasks \cite{Goyal2019ws,ericsson2021well}. 
As a result of not having access to any semantic supervision (\eg discrete category labels in the case of image classification), current best performing self-supervised methods typically use aggressive image augmentation strategies to generate different ``views'' of an input image during training \cite{wu2018unsupervised,he2020momentum,chen2020simple,grill2020bootstrap,chen2021exploring}. 
The training time objective then consists of pulling these distinct ``views'' of the same image close to one another in feature space. 
The ability to generate plausible image variation using these manually designed image augmentation strategies would then appear to be a limiting factor in the further advancement of self-supervised learning (SSL).

\begin{figure}[t]
\begin{center}
 \includegraphics[width=0.9\columnwidth]{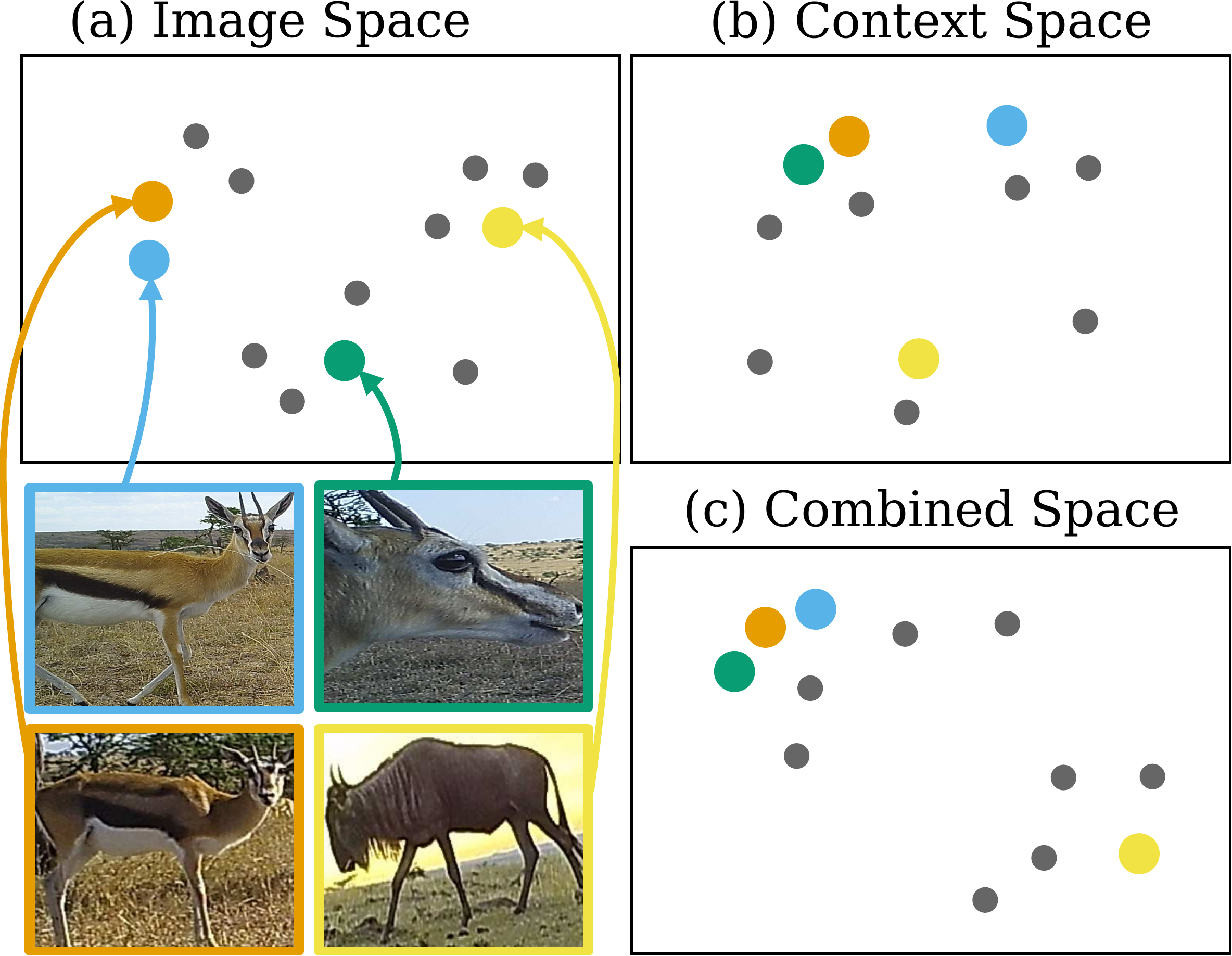}
\end{center}
   \vspace{-10pt}
   \caption{(a) Conventional self-supervised methods are capable of bringing visually similar examples closer in image embedding space.
   (b) There is often rich context information (\eg encoding where and when an image was captured) that can also convey similarity. 
   (c) By combing the complimentary nature of both signals, we can improve the quality of the final embedding space.
   }
   \vspace{-15pt}
\label{fig:overview}
\end{figure}

Current state-of-the-art self-supervised methods have predominantly been designed using image collections originally constructed for \emph{supervised} learning \eg \cite{deng2009imagenet,zhou2017places}. 
This has necessitated the exploration of different augmentation strategies to introduce appearance variation during training. 
However, a more natural signal to use is to exploit the fact that image observations that are made close in time and space are very likely to contain the same object instance. 
This form of natural variation has been used in self-supervised learning from video \cite{sermanet2018time,purushwalkam2020demystifying,qian2021spatiotemporal} or spatially distributed image collections \cite{jean2019tile2vec,ayush2020geography}. 
More generally, one can think of having access to not only a collection of images at training time, but also potentially rich context information pertaining to when and where each image was captured, in addition to other cues.

The central question that we address in this paper is how to make use of this context information during self-supervised learning to select more useful and varied image pairs at training time. 
The aim is to provide the self-supervised algorithm with ``natural'' positive image pairs and thus to establish connections in the latent feature space that were not possible with conventional augmentation-based approaches (Figure~\ref{fig:overview}). 
We evaluate several different approaches and show, perhaps surprisingly, that the choice of images has more of an impact on performance than the underlying self-supervision algorithm.
Our analysis is applicable to any self-supervised method that attempts to maximize the similarity between two ``views'' of the same visual concept, even those that do not require negative sample pairs~\cite{chen2021exploring}.

We focus our evaluation on image collections that have been captured using camera traps - also known as ``wild-cams'' or ``trail cameras''. 
These types of images are commonly collected for the purpose of biodiversity monitoring~\cite{yu2013automated,norouzzadeh2018automatically,beery2018recognition,tabak2019machine,beery2020context,glover2019camera}.  
Unlike more conventional image datasets typically used by the computer vision community, camera trap images exhibit some interesting properties that make them particularly well suited to evaluating self-supervised learning:  
(i) Camera trap images are not captured by humans directly, instead the cameras are automatically triggered based on the proximity of nearby animals.  
This overcomes the ``iconic'' object view bias that tends to be prevalent in datasets like ImageNet~\cite{deng2009imagenet} or iNaturalist~\cite{van2018inaturalist,van2021benchmarking}. 
They also contain other challenges such as partial depictions of objects due to occlusion, significant scene illumination changes, and strong object and location correlations~\cite{beery2018recognition};
(ii) The images are often captured in short temporal bursts and come with rich context data in the form of the time of day, the time of year, and location, and may also include additional information characterizing the local habitat. 
This information can provide useful clues about what animal species are potentially present at a given location; 
(iii) A conservative estimate is that tens of thousands of ever-active cameras are deployed globally~\cite{steenweg2017scaling}, necessitating large amounts of tedious work on the part of ecologists and conservation biologists to manually annotate the incoming images or to correct errors made by automatically generated classifier predictions.  
Informative image features, derived from self-supervised learning, could significantly reduce this manual effort and have the potential to be an important tool in aiding the critical task of scalable global biodiversity monitoring. 

We make the following three contributions: 
\vspace{-6pt}
\begin{enumerate}
    \setlength\itemsep{-4pt} 

    \item We explore the benefits of self-supervised learning on four challenging camera trap datasets. We observe that self-supervised features are significantly more effective on average for downstream classification compared to widely adopted transfer learning baselines.
    
    \item We show that how images from these datasets are selected during self-supervised training  has much more impact on the quality of learned features compared to the choice of the underlying self-supervised training loss that is used. 

    \item While the role of \emph{negative} image pairs in self-supervised learning has received significant attention, we show that current methods are surprisingly robust to incorrectly selected \emph{positive} image pairs during training. 
    This provides an important insight for the design of future self-supervised methods.
    
\end{enumerate}

\section{Related Work}

\subsection{Self-Supervised Learning}
\vspace{-5pt}
The goal of self-supervised learning (SSL) from visual data is to learn a function that can extract semantically meaningful image representations, without requiring any semantic annotations at training time. 
Until recently, the main focus of SSL research in vision centered on designing proxy tasks that, when solved at training time, would result in useful features for downstream transfer learning. 
Examples of approaches include visual puzzle solving \cite{noroozi2016unsupervised}, image colorization \cite{zhang2016colorful,zhang2017split}, image in-painting~\cite{pathak2016context}, and image rotation prediction \cite{gidaris2018unsupervised}, to name a few. 

There has been recent excitement for contrastive-based approaches~\cite{hadsell2006dimensionality,gutmann2010noise,oord2018representation}. 
Instead of requiring elaborate proxy tasks, contrastive methods learn features by pushing positive image pairs close to each other in feature space, while pushing negative pairs away from each other. 
Given semantic information~\cite{schroff2015facenet,khosla2020supervised}, positive and negative pairs can be defined based on object category labels, or some other form of semantic supervision. 
However, for SSL, this information is not available. 
As a result, the current most common solution is to artificially augment images with random transformations~\cite{dosovitskiy2014discriminative} to create positive and negative image pairs. 
Specifically, positive pairs are differently augmented versions of the same image, while negative pairs are augmented versions of any two \emph{different} images. 
The assumption is that the selected augmentation space preserves semantic content, while introducing noise that the feature extractor will learn to tolerate through invariance.

Contrastive-based methods require negative image pairs, and various strategies have been proposed to ensure that useful negatives (\ie hard negatives) are selected during training. 
Common strategies to ensure challenging negatives are selected include non-parametric memory banks~\cite{wu2018unsupervised,zhuang2019local}, 
momentum encoders \cite{he2020momentum}, auxiliary modalities~\cite{tian2020contrastive}, local region rather than full image reasoning~\cite{oord2018representation,bachman2019learning}, learning data prototypes  online~\cite{caron2020unsupervised}, or simply training with large batch sizes~\cite{chen2020simple}. 
The body of work in this space is growing fast, see~\cite{schmarje2021survey} for an overview.

One of the typical assumptions made by conventional contrastive-based approaches, when constructing negative pairs, is that randomly sampled images from the same batch, or from a non-parametric memory, do not depict the same semantic content. In practice however this assumption is often violated, and thus results in a large number of false negative pairs during training. 
Recent research~\cite{chuang2020debiased} has quantified the impact of this assumption by reporting performance gains under the presence of oracle (\ie ground truth) negative labels  and proposed an unsupervised negative sampling technique that partially improves performance by correcting for the introduced error. 
Another study~\cite{robinson2020contrastive} also observed the importance of hard negatives during SSL and introduced a user controllable hard negative sampling approach that improves over the baseline of uniform sampling.
\cite{kalantidis2020hard} showed that hard negatives are required for effective SSL and proposed an unsupervised feature space mixing method inspired by~\cite{zhang2017mixup}, to generate challenging training examples.

Given the aforementioned obstacles associated with effective negative sample selection, there have been recent attempts to forego the need for negative image pairs during SSL. 
One set of methods involve using additional positive prediction-based networks~\cite{grill2020bootstrap} or layers~\cite{chen2021exploring} during training.
These approaches compete with contrastive methods, and have the added benefit of not requiring large batch sizes as they do not need to sample from a large pool of negatives.
In light of this diminished reliance on negatives, we turn our attention to the role of the \emph{positive} image pairs. 
We show that, perhaps unsurprisingly, there is a large performance gap between standard SSL methods and oracle baselines (\ie given ground truth labels).  
By making use of context information, we can select positive pairs that exhibit more visual diversity compared to standard augmentation resulting in improved downstream performance.

\begin{figure*}[ht]
\begin{center}
 \includegraphics[width=0.9\linewidth]{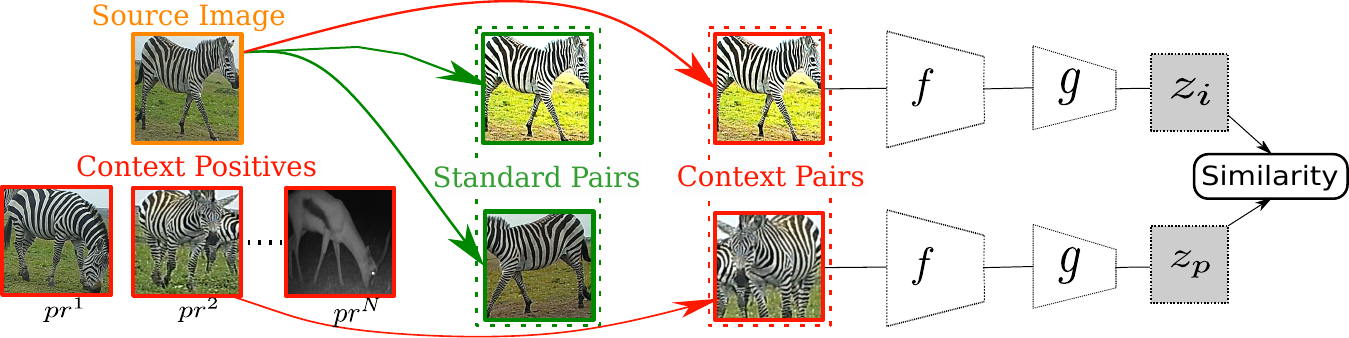}
\end{center}
   \vspace{-10pt}
   \caption{Overview of our approach. 
   Conventional SSL typically generates augmented pairs of an input source image. Those are then passed to the model (right). 
   Instead, we advocate that the careful selection of image pairs based on their context similarity (\eg in space or time), generates more varied and useful image pairs that result in more informative visual representations.
   }
\label{fig:method_plot}
\vspace{-15pt}
\end{figure*}

\subsection{Context-Aware Classification}
\vspace{-5pt}
While there are notable exceptions, for the most part, the computer vision community has predominately focused on using benchmark image datasets that lack additional data beyond the images themselves and their task-specific supervised annotations \eg~\cite{deng2009imagenet,lin2014microsoft}. 
Images acquired in the wild, as opposed to scraped from the web, often come with valuable metadata, encoding where and when the images were captured, along with other potentially valuable cues.   
We broadly refer to this free weak supervision as \emph{context} data.  

There are many instances where specific visual concepts can only be disambiguated by knowing where the images were captured.  
For example, in case of species identification, some categories may only be correctly classified when the image location is known at test time. 
Existing supervised works have explored the use of this geographical context via non-parametric density estimation to model the spatial distribution of object categories~\cite{berg2014birdsnap}. 
Others use deep networks with integrated geographical encoding layers~\cite{tang2015improving,chu2019geo}, or factorized models that reason about object spatial distributions in a latent feature space~\cite{mac2019presence}. 
The previously mentioned approaches are all supervised, but geographical proximity has also been successfully used as a signal for SSL. 
In the case of representation learning from aerial images, recent analyses \cite{ayush2020geography,jean2019tile2vec} make the assumption that geographically close-by locations should be more similar in their latent features compared to more distant locations, and show that this results in superior image features. 

Another form of readily accessible contextual information are temporal cues. 
For short temporal changes (\eg nearby frames in a video), the observed raw pixel data might vary significantly from frame to frame, but it is often reasonable to assume that the underlying latent content being observed changes more slowly~\cite{wiskott2002slow}. 
The assumption that nearby frames contain the same object instance has previously been successfully exploited in camera trap images by merging predictions from nearby time steps at test time~\cite{beery2018recognition}. 
Long range temporal reasoning is also a common component of supervised video analysis~\cite{wu2019long,feichtenhofer2019slowfast}, but without supervision it is more challenging to make use of this information. 
Combined spatial (in image space) and temporal reasoning was recently shown to be effective for the problem of object detection in static camera traps~\cite{beery2020context}. 
By exploiting the fact that objects (\ie people or animals) may exhibit similar behavior over time at the same location, the proposed attention-based mechanism was able to use this signal to improve detection performance at test time, where no supervision is available. 
However, supervision in the form of bounding boxes and ground truth category labels are still required at training time. 

Several approaches have been proposed for SSL in video, including posing the learning problem as a prediction task \eg predicting future events~\cite{srivastava2015unsupervised}, predicting motion and appearance statistics \cite{wang2019self}, predicting sequence order \cite{fernando2017self,lee2017unsupervised}, and also contrastive learning~\cite{qian2021spatiotemporal}.
While the camera trap datasets that we use are related to video datasets, in that they also contain some temporal information, the frames are not uniformly sampled over time, but instead, the cameras are trigger-based events in the scene. 

In this work, we evaluate different mechanisms for selecting positive image pairs during SSL. 
We also present a context-based image selection approach for choosing high probability positive pairs at training time to learn more effective self-supervised visual features. 
To achieve this, we make use of both spatial and temporal context in an unsupervised fashion.
\section{Method}

In self-supervised visual representation learning, we assume that we have a set of images $x_i \in \mathcal{X}$ at training time, but do not have access to any associated ground truth supervision \eg object category labels. 
Our goal is to learn the parameters of a feature extractor $f$, so that when we apply the network on an image $x_i$, we obtain a feature vector that can be used for downstream tasks. 
Alternatively, we may also wish to fine-tune the feature extractor in an end-to-end fashion with additional supervision under the assumption that this will serve as a better initialization than randomly initialized parameters. 
In the case of images, the feature extractor $f$ can be parameterized as a deep convolutional neural network, \eg a deep Residual Network~\cite{he2016deep}. 

To overcome the lack of supervision at training time, many recent approaches rely on variants of augmentation-based SSL in an attempt to learn expressive features from the raw image data. 
Specifically, the standard pipeline involves taking a source image $x_i$ and performing two sets of synthetic augmentations on the image (\eg cropping, flipping, color jittering, \etc) to create two alternative ``views'' of the source image $(\tilde{x}_i, \tilde{x}_p)$. 
See \cite{chen2020simple} for representative augmentations. 
We refer to these two images as a positive augmented pair. 
The goal of the augmentations is to introduce visual diversity, while ideally preserving the underlying semantic information contained in the source image. 
During self-supervised training, the learning algorithm attempts to push the positive pair of images close to each other in feature space. 
The intuition is that we want to learn a feature extractor $f$ that  projects images containing the same visual concepts to similar regions in feature space. 
Later, we describe different approaches for selecting varied positive pairs. 
An overview of our pipeline is shown in Figure~\ref{fig:method_plot}. 

\subsection{Self-Supervised Losses} \label{sslapproaches}
Here we outline three broad methods for SSL that serve as the basis for our experimental evaluation. 
Common to each method is that they take the output of the feature extractor $f$ and pass it through a projection network $g$, resulting in a lower-dimensional embedding vector $z_i = g(f(\tilde{x}_i))$. 
This projection network is commonly represented as a fully connected multi-layer neural network~\cite{chen2020simple}. 

\noindent{\bf Triplet}~\cite{weinberger2009distance,schroff2015facenet}.
The conventional triplet-based margin loss aims to push positive embedding pairs  $(z_i,z_p)$ close to each other, while also pushing negative pairs $(z_i,z_n)$ further away from each other. 
Here, a ``negative'' pair, is a pair derived from two different images $(\tilde{x}_i, \tilde{x}_n)$, where $i \neq n$.
In the language of triplet-based learning, we want the distance between the anchor $z_i$ and the positive $z_p$ to be smaller than the distance between the anchor and the negative $z_n$.
The triplet loss is defined as
\begin{equation}
L^T_{i}= \max\left(D(z_i,z_p) - D(z_i,z_n) + m, 0\right),
\end{equation}
where $m$ is a scalar representing the desired margin of the loss, and $D(.)$ is an appropriate distance function, \eg Euclidean distance or negative cosine similarity. 
The triplet loss requires positive pairs and, crucially, negative pairs. 
For SSL, we can generate positive pairs through image augmentation, and a common strategy for selecting negative pairs, in the absence of any other knowledge, is to simply randomly sample them from the entire dataset. 
While not commonly used for SSL, later we show that the triplet loss is surprisingly effective when compared to recent methods. 

\noindent{\bf SimCLR}~\cite{chen2020simple}.
Building on related approaches \cite{sohn2016improved,wu2018unsupervised}, at the core of SimCLR is the use of a normalized temperature-based softmax cross-entropy loss.  
As with the triplet loss, the aim is to minimize the positive pair distance, while maximizing the negative pair distance. 
However, unlike the triplet loss, where we select one negative pair for each positive one, here we select a set of negatives. 
Unlike methods that maintain a memory bank containing embeddings for the entire training set to select negatives from \cite{wu2018unsupervised}, SimCLR simply uses large training batches and samples the negatives for each source instance from within the same batch. 
The loss used in~\cite{chen2020simple} over a set of items $\mathcal{B}$ is defined as 
\begin{equation}
 L^S_{i}= - \log \left( \frac{\exp({-D(z_i,z_p)/ \tau})}{\sum_{n \in \mathcal{B}}\mathds{1}_{[n\neq i]} exp({-D(z_i,z_n)/ \tau}) } \right),
\end{equation}
where $\tau$ is a temperature parameter used to control the scale of the distances and $\mathds{1}_{[n\neq i]}$ ensures that an embedding vector is not compared with itself. 
Importantly, here the source-positive pair is also part of the sum in the denominator to ensure that the loss is normalized correctly \ie $p \in \mathcal{B}$. 
 
\noindent{\bf SimSiam}~\cite{chen2021exploring}.
SimSiam is a non-contrastive self-supervised method that innovates on the previous two losses by removing the need for negative pairs during training. 
This is an important simplification because, as stated earlier, choosing appropriate negative samples for SSL is challenging. 
\cite{chen2021exploring} introduced an additional fully connected network $h$, termed the prediction network. 
The goal of this network is to predict the embedding vector of the other image in the input pair.
To avoid collapsing to a degenerate solution in the absence of negative examples, SimSiam applies a bottleneck structure in the prediction network and a stop-gradient operation~\cite{grill2020bootstrap} to prevent information flowing back through the projection and feature networks for a given image in the input pair. 
During training, the goal is to maximize agreement between the embedding vector $z_i$ and the prediction $h(z_p)$ using the following symmetrized loss
\begin{equation}
L^M_{i}= \frac{1}{2} D(h(z_i),\texttt{sg}(z_p)) + \frac{1}{2} D(h(z_p),\texttt{sg}(z_i)).
\end{equation}
Here, $\texttt{sg}$ represents the application of the stop-gradient. 
Despite not requiring any negative image pairs, SimSiam still performs competitively compared to contrastive approaches, even when using more modest batch sizes~\cite{chen2021exploring}. 

\subsection{Selecting Positive Image Pairs} \label{positivemining}
One commonality shared by the three previous self-supervised methods is that they each require positive image pairs during training \ie $(x_i,x_p)$. 
Given only images, there are limited options for how to create these positive pairs. 
However, when additional information, such as time or location is available, this opens the door to alternative methods for selecting training pairs. 

\noindent{\bf Image Augmentations}.
The standard approach used by many SSL methods that only train on unordered image collections is to create positive image pairs by stochastically augmenting the source images during training \ie $(\tilde{x}_i, \tilde{x}_p)$, where $i=p$.
Note that we also apply these augmentations to each of the following selection mechanisms, even if the image pairs are different, \ie when $i \neq p$.

\noindent{\bf Oracle Positive Selection}.
If category labels were available, one could use them  to select positive image pairs~\cite{khosla2020supervised}.
Here, we assume that our input image collection consists of pairs of $(x_i, y_i)$, where $y_i$ is the corresponding category label for image $x_i$. 
For each image $x_i$, $x_p$ is selected from the set of images belonging to the same category \ie $y_i = y_p$.  
This represents an idealized setting, and is not practical for actual SSL where this supervision is absent. 
However, it provides useful insight, and gives us an upper bound on performance. 
For methods that also require negative pairs, one could also select them using a similar procedure~\cite{chuang2020debiased}, \ie $y_i \neq y_n$. 
Our focus is on the role of positive images, and as a result, we select negative pairs using random sampling in all cases. 
Later, we explore variants of this oracle positive selection mechanism where we introduce noise, so that with some probability $\lambda$, we intentionally select $p$ to be from a different category, \ie $y_i \neq y_p$. 

\noindent{\bf Sequence Positives}.
Given sequential input data \eg a video or burst of frames, one strategy is to select positive pairs based on their proximity in time. 
Here we choose images as pairs if they are from the same location and are within a specified number of frames or a unit of time from each other. 
We also ensure that the source image itself can be selected.  
This has the advantage of introducing more natural visual diversity compared to what is possible from conventional image-based augmentation. 
The disadvantage of this approach is that one must commit to a hard threshold on the number of frames, or unit of time, when deciding what constitutes a potential positive pair. 

\noindent{\bf Context-Based Selection}.
In many real world datasets (\eg camera traps) there can be extra information available in addition to the raw images. 
We can assume that each image $x_i$ is associated with a $K$ dimensional context vector $c_i$. 
This vector could include information such as an encoding of the geographical location of the image, date, time of day, the pixel coordinates of the image (\ie it could be cropped from a larger image), \etc.  
Instead of having to specify thresholds for each of these different dimensions, a simpler approach is to define a distance measure between a pair $i$ and $j$, 
\begin{equation}
    D^c(c_i,c_j)=  \sqrt{{\sum_{k=1}^{K} (c^k_i-c^k_j)^2  }}.
\end{equation}
When selecting a positive pair at training time for a source image $i$, we simply construct a distribution over all pairs 
\begin{equation}
    pr_{i}^{j} = \frac{\exp(-D^c(c_i,c_j)/\tau_c)}{\sum_{n=1}^N\exp(-D^c(c_i,c_n)/\tau_c)}.
\end{equation}
Here, $N$ is the number of images in the unlabeled training set, which includes $i$ and $j$, and $\tau_c$ is another temperature hyperparameter. 
Then to select a positive item $p$ for each $i$, we sample from a categorical distribution parameterized by the vector $pr_{i}$, \ie $p \sim \text{Categorical}(pr_{i})$.
As a result, items that are close in context space will have a higher probability of being selected as a positive pair. 
Note that there is also a chance that we sample the same image as the source, \ie $p=i$. 
As we perform random augmentations on each view of the source image, in cases where $p=i$, this results in the standard image augmentation-based selection mechanism. 
\section{Experiments}

\subsection{Datasets}
\vspace{-5pt}
We perform experiments on four representative camera trap datasets below and one satellite dataset~\cite{christie2018functional} in the supplementary. 
The camera trap datasets exhibit complementary properties, \eg collected in multiple countries versus one region, or collected over long (years) versus shorter (months) time periods. 
Camera trap images contain many difficulties that make automatic identification particularly challenging~\cite{beery2018recognition}.
While other benchmark datasets contain some useful context data related to when and where the images were taken \eg \cite{tang2015improving,van2018inaturalist,van2021benchmarking}, they do not explicitly contain sequences of images captured over time from the same location.
We assume that the images are manually cropped around the objects of interest, since we are concerned with the problem of classification, not detection. 
High precision detectors are available, specifically tuned for camera trap images, in cases where ground truth bounding boxes are unavailable~\cite{beery2019efficient}.
Image counts below refer to cropped images.
The test splits for all datasets, except Snapshot Serengeti, were generated according to the protocol in~\cite{beery2018recognition}. 
This is designed to evaluate generalization behavior across novel camera trap sites. 
Each dataset consists of a set of locations (\ie camera trap deployments), with multiple images from each location. 
Some locations are shared between the train and test sets and some are only present in the test set.

\noindent{\bf Caltech Camera Traps} (CCT20) \cite{beery2018recognition}.   
CCT20 contains images from 20 different sites in the Southwestern United States.  
In total, the dataset includes 15 species such as rabbits, raccoons, coyotes, bobcats, and others. 
There are 10 distinct camera trap locations in the test set and 10 that are shared between train and test, resulting in 12,617, 3,436, and 32,050 images in train, validation, and test.

\noindent{\bf Island Conservation Camera Traps} (ICCT) \cite{islandconservation2020}. 
ICCT consists of images taken across seven different islands, spanning six countries. 
Due to the global distribution of the dataset, it contains diverse ecosystems such as tropical, dry, and temperate. 
There are 116 locations, with 56 observed for training. 
We retain 12 species that had at least 100 appearances in the training locations, resulting in 11,378, 1,684, and 41,527 images in train, validation, and test.

\noindent{\bf Snapshot Serengeti} (Serengeti) \cite{swanson2015snapshot}. 
This contains cameras located around the Serengeti National Park in Tanzania, collected over multiple seasons. 
We use the subset that includes bounding box annotations.
We use the author-provided train and test split, and retain at most 1,000 full images (\ie not crops) for each species, keeping species that have at least 100 appearances. 
This results in 39 species, with 32,702 train and 8,492 test cropped images, from 179 train and 45 test locations. 
In contrast to the other datasets, the train and test locations do not overlap.

\noindent{\bf Maasai Mara Camera Traps} (MMCT). 
Our final dataset 
is a fixed subset of our own ongoing collection efforts, which we will make available. 
MMCT contains images from 176 sites across the Maasai Mara in Kenya, which is primarily an open savanna and is known for its rich wildlife. 
It shares some ecosystem characteristics with Snapshot Serengeti, but it exhibits a larger class imbalance and more domesticated animals, given that the study was designed to study human impact on the environment. 
It contains 20 species of mammals, ranging from wildebeests, elephants, through to giraffes, and includes many threatened species. 
In total, there are 10,243 train images from 85 locations, along with 3,306 validation and 31,841 test images. 

\subsection{Implementation Details} 
\vspace{-5pt}
For all experiments, we use a ResNet18~\cite{he2016deep} as our feature extractor and add a two layer MLP for the projector $g$.
For SimSiam, we use an additional MLP for the predictor $h$. 
As our input images are already cropped around the object of interest, we train with images of size $112\times112$. 
During self-supervised training, we augment the images with a set of transformations, similar to those used in SimCLR \cite{chen2020simple}.
Unlike existing SSL methods that train from randomly initialized weights, we start from ImageNet pretrained weights and train for an additional 200 epochs with a batch size of 256. 
Results with random initialization are available in the supplementary material. 
A minor contribution is showing that SSL is still effective in this regime. 

For each individual camera trap image, we construct a corresponding context vector $c$. 
While many different context variables are possible, for now we restrict ourselves to those encoding time and location.
We encode date and time using wrapped coordinates as in~\cite{mac2019presence}. 
For many camera trap datasets, the GPS coordinates of each camera are often kept secret as some of the animals present many be endangered.
To overcome this, we use a one hot encoding to represent the deployment ID. 

To evaluate the effectiveness of the different SSL approaches, we use the standard linear evaluation protocol \cite{zhang2016colorful,oord2018representation,chen2020simple}. 
Specifically, after self-supervised training we only keep the backbone feature extractor $f$, and use it to compute features for the training and test images. 
We then train a linear classifier using these features, and evaluate performance across the low to high data regimes \ie by using 1\%, 10\%, or 100\% of the available training labels. 
Unlike standard computer vision datasets~\eg~\cite{deng2009imagenet}, camera trap datasets tend to be heavily imbalanced, making this a particularly challenging task when there are few labels. 
Additional implementation details, hyperparameters, and experiments can be found in our supplementary material. 

\subsection{Results} 
\vspace{-5pt}

\begin{figure}[t]
\begin{center}
 \includegraphics[width=0.88\columnwidth]{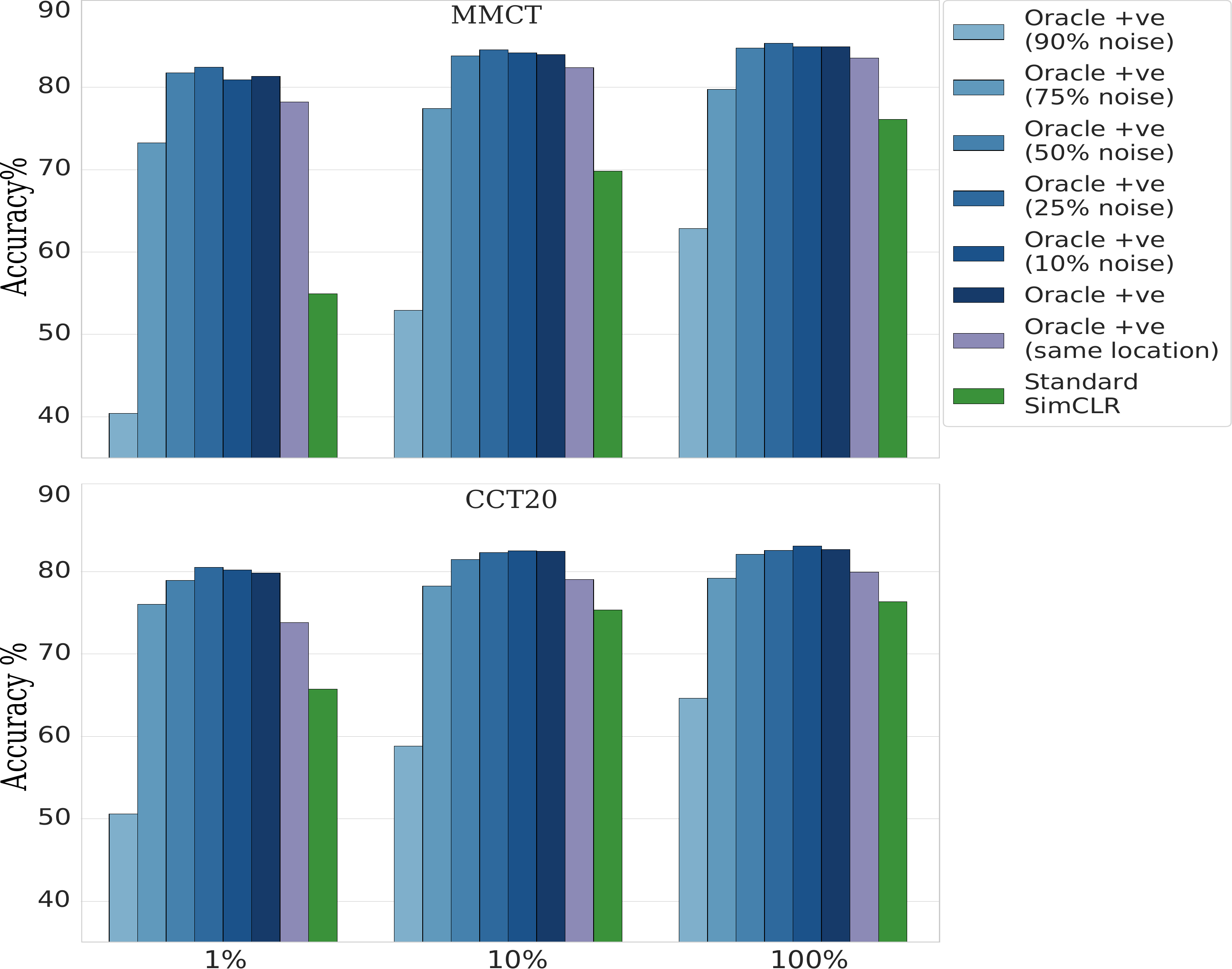}
\end{center}
   \vspace{-5pt}
   \caption{Oracle positive pair selection with SimCLR. 
   We observe that SimCLR is surprisingly robust to label noise in the oracle positive selection setting. 
   Performance only significantly deteriorates when the amount of noise is 90\%. 
   1\%, 10\%, and 100\% refer to proportions of supervised labeling used for evaluation after SSL.
   }
   \vspace{-15pt}
\label{fig:pos_results}
\end{figure}

\noindent{\bf Oracle Positive Selection}. \label{pos_importance}
We first validate our hypothesis that SSL on camera traps data can benefit from seeing natural variation in the positive pairs when compared to standard image augmentations. 
We consider an idealized scenario where an oracle has access to the ground truth species labels during self-supervised training. 
We then sample positive pairs using the oracle selection strategy outlined in Section~\ref{sslapproaches}. 
In Figure~\ref{fig:pos_results}, we observe a significant gap between standard SimCLR, which generates image pairs by augmenting the source image alone, and an oracle variant of SimCLR (``Oracle +ve'') that selects true positives based on the ground truth species labels. 
More specifically, for MMCT, the top-one accuracy difference ranges from  $\sim$9\% and $\sim$28\%,  and between $\sim$6\% and $\sim$15\% for CCT20 across the different amounts of downstream supervision.
It is worth noting that the oracle selection mechanism can select pairs from across the entire training set (\ie from different locations and different points in times). 
We also include an oracle baseline that can only select positive pairs from the same camera trap deployment (\ie the same physical location but the time could vary by as much as months). 
``Oracle +ve (same location)'' is also significantly better than standard SimCLR. 
These results are encouraging as it provides evidence that more sophisticated methods for grouping images based on context from the same deployment (\eg \cite{beery2020context}) could result in further performance improvements.

\noindent{\bf Robustness to Oracle Noise}. 
The oracle positive baseline in the previous section is an unreasonably strong and unrealistic baseline. 
We also evaluate the case where there is noise associated with the positive pair selection. 
Specifically, in Figure~\ref{fig:pos_results}, ``Oracle +ve (X\% noise)'' indicates that X\% of the time, the oracle makes a mistake, pairing two images from different classes, \ie $y_i \neq y_p$. 
Perhaps surprisingly, we observe across both datasets, that SimCLR is quite robust to this error, at least until the noise percentage gets very large \eg 90\%. 
Clearly, there is a benefit from being able to select images pairs that exhibit visual variety that would not be possible to create through image augmentation alone. 
This motivates our objective of selecting more varied positives by means of context-based mining. 

\begin{figure}[h]
\begin{center}
 \includegraphics[width=0.9\columnwidth]{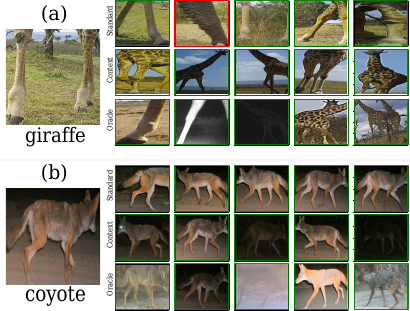}
\end{center}
   \vspace{-5pt}
   \caption{Here we show the top five nearest neighbors from the test set in the 128-dimensional embedding space (\ie the output of the projector) from (a) MMCT and (b) CCT20, for three SimCLR variations. We see that the nearest neighbors for `Standard' SimCLR display limited visual diversity. The unobtainable `Oracle' model, which has been trained with ground truth labels, has the most variety. Our `Context' approach is between the two extremes and shows non-trivial diversity which indicates that it contains more semantic information in its features compared to the standard augmentation-based SimCLR. Green and red outlines correspond to similar and different class neighbors respectively.}
   \vspace{-15pt}
\label{fig:synth_pairs}
\end{figure}

\begin{figure*}[t] 
\begin{center}
 \includegraphics[width=0.9\linewidth,keepaspectratio]{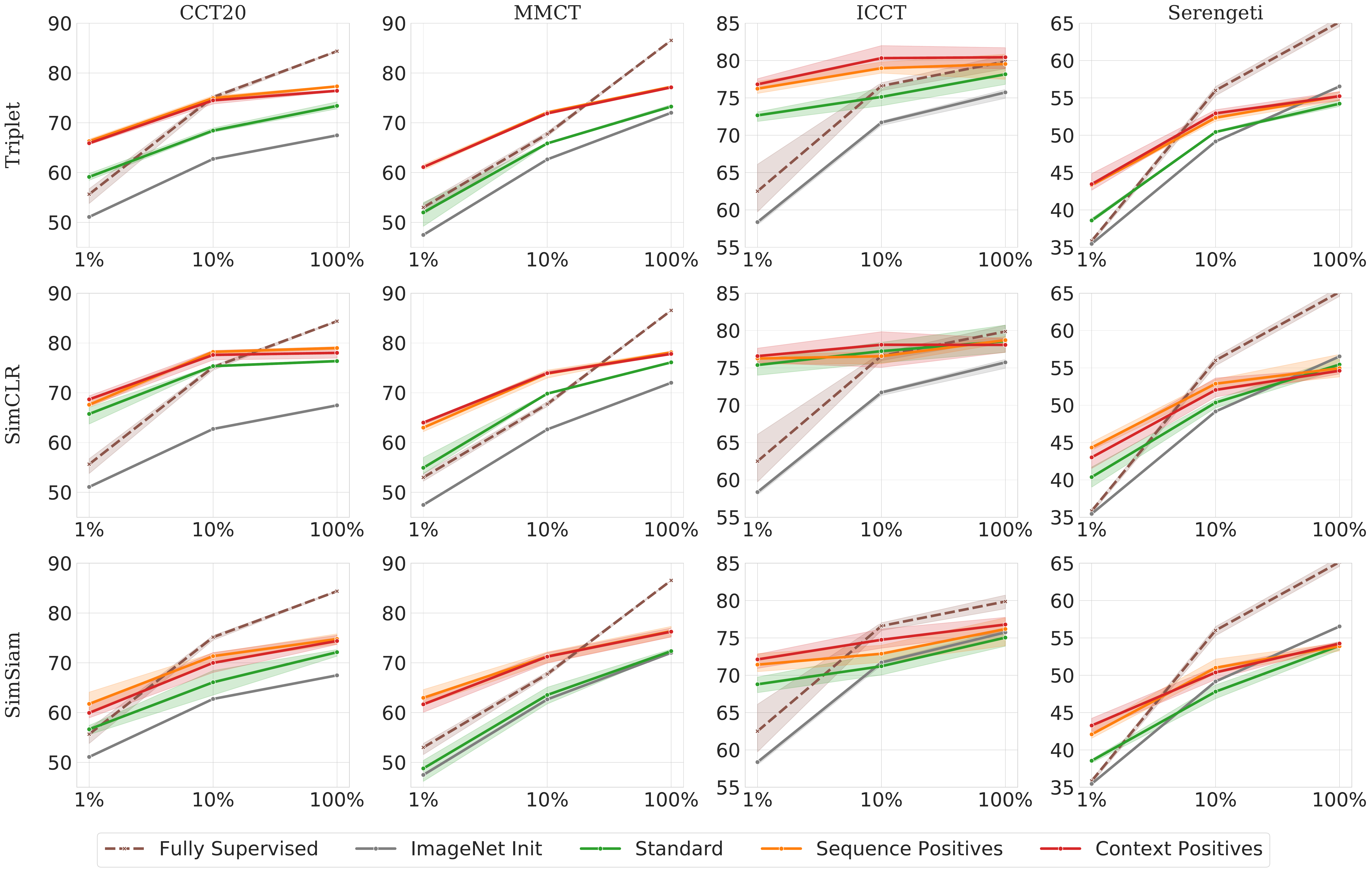}
\end{center}
   \vspace{-10pt}
   \caption{
   Top-one classification accuracy across four different camera trap datasets (columns) and three different SSL algorithms (rows). 
   Each line represents a different mechanism for selecting positive image pairs during self-supervised training. 
   Accuracy is computed by evaluating linear classifiers trained on different amounts of supervision (\ie 1, 10, or 100\%) from the respective training sets. The error bars represent the average accuracy and standard deviation  over three repeats of SSL. 
   For reference, we also include fully supervised baselines, trained end-to-end, with the proportion of labels indicated on each x axis. 
   We see that SSL methods significantly outperform ImageNet derived features in nearly all cases. Furthermore, using natural positives (Sequence Positives or Context Positives) is always better than standard image augmentation alone (Standard), and has more impact than the choice of SSL method. 
   }
\label{fig:main_results}
\end{figure*}

\noindent{\bf Impact of Positive Selection}.
In Figure~\ref{fig:main_results} we present an extensive set of experiments across four different camera trap datasets, three SSL methods, and three positive pair selection approaches. 
The first observation is that standard SSL results in superior representations for downstream linear classification compared to standard ImageNet initialized features in nearly all settings (``Standard'' versus ``ImageNet Init''). 
The second observation is that training with more ``natural'' positives is superior to standard SSL in almost all cases (``Sequence Positives'' or ``Context Positives'' versus ``Standard'').
In many instances, this performance difference can be $>5\%$ top one accuracy. 
Our ``Context Positives'' approach is consistently ranked first or second (excluding end-to-end supervised training which is provided for reference). We speculate that further gains may be achieved from using richer context information. 
Figure~\ref{fig:synth_pairs} shows examples of the nearest neighbors retrieved from the different models. 
We can see that ``Context Positives'' results in more visual diversity compared to standard SSL.

\noindent{\bf Impact of Self-Supervised Algorithm}.
One important observation from Figure~\ref{fig:main_results} is that the choice of positive selection mechanism has more impact on downstream classification accuracy than the actual self-supervised method used. 
In fact, we show that even the conventional triplet loss when combined with any type of context information, results in better performance compared to recent state-of-the-art methods like SimCLR or SimSiam, when they only use standard image augmentation. 
This indicates that for ``in the wild'' datasets like camera traps, additional attention should be given to alternative approaches for positive image pair selection. 
This is further motivated by the large performance gap that still exists between the self-supervised methods and the oracle baselines in Figure~\ref{fig:pos_results}. Moreover, we observe that between the three self-supervised approaches, given the proposed positive pair selection, SimCLR performs slightly better on average.

\noindent{\bf Discussion and Limitations}.
The datasets used for our evaluation have far fewer classes compared to datasets such as ImageNet~\cite{deng2009imagenet} or iNaturalist~\cite{van2018inaturalist,van2021benchmarking}. 
However, we argue that they are in fact more representative of the thousands of medium scale datasets that are generated by biodiversity researchers around the world every month. 
We also assumed that our images are cropped around the target of interest. 
While camera trap datasets can be very large, the majority of the images are often empty.
Pretrained object detectors~\cite{beery2019efficient} can be used to filter the images to only these boxes of interest (\ie animals). 
We present results with pretrained detectors in the supplementary material. 
Currently, our context-based sampling approach assumes that all context dimensions are weighted equally. 
Given how effective we have shown context information to be for self-supervised learning, a natural question is how can we improve performance further \eg via learned context weights. 
Without supervision, this is a challenging problem which we leave for future work. 

\section{Conclusion} 
We have explored the problem of self-supervised learning in camera trap datasets. 
We showed that these types of image collections are a valuable testbed for benchmarking advances in self-supervised learning as they are  complementary to the standard benchmarks commonly used in vision. 
In fact, conventional wisdom from well-explored standard benchmarks does not transfer directly to these types of images, as we observe that how images are selected during training can have a larger impact than the specific self-supervised algorithm. 
We posit that further exploration of this setting will lead to a greater understanding of the potential for self-supervised learning for ``in the wild'' collected datasets, and importantly, will directly benefit critical applications such as biodiversity monitoring. 

\vspace{5pt}
\noindent{\bf Acknowledgements.}
The MMCT images are collected as part of the Biome Health Project \cite{biomehealth} funded by WWF-UK. Thanks to Holly Pringle, Emily Madsen, Alex Rabeau, and Georgia Cronshaw for the annotation of the MMCT data.

\clearpage
{\small
\bibliographystyle{ieee_fullname}
\bibliography{main.bib}
}

\clearpage
\appendix

\section{Additional Experiments}
Here we provide additional experiments that highlight the success and failure cases of the different self-supervised learning (SSL) algorithms and our positive image selection mechanisms we evaluated in the main paper. 

\subsection{Are modern SSL algorithms effective on camera trap data?}
From Figure~5 in the main paper we can see that self-supervised methods are superior to ImageNet derived features, in the vast majority of cases. In addition, context-based approaches for positive image sampling further increase the performance. 
In Figure~\ref{fig:method_comparison_results}, we can see that while SimCLR is typically the best performing method, there are multiple instances where more naive SSL methods outperform it. 
Using context-based sampling increases overall accuracy (bottom \vs top row), making the simpler baselines competitive with conventional SimCLR.  

\subsection{Do we need ground truth bounding boxes for SSL to be effective for camera trap data?}

In the main paper, we performed all experiments using ground truth bounding boxes. 
This was to avoid drawing conclusions based on any biases that may be present in a specific detector \eg some of the datasets we use are public and thus could have been part of the training set for a public detector. 
To evaluate how effective detected boxes are, we replaced ground truth detections with automatically derived ones from MegaDetector (MD)~\cite{beery2019efficient} and retrained our SSL models from scratch on the MMCT dataset. 
We used MMCT because we can guarantee that MegaDetector is not trained on it.
To ensure a fair comparison with the results in the main paper, when performing linear evaluation (but not when performing SSL), we used the ground truth boxes for the test set and supervised training set during evaluation. 
We can see from the results in Table~\ref{tab:mega_detector} that the quality of representations from detected boxes are comparable to the ones learned from ground truth (GT) boxes. 

Next, we increased the number of detected boxes by a factor of two by adding more images to the training set \ie from the same camera locations, but at different points in time (denoted as MDx2 v1). 
This resulted in about 25,406 total cropped images, as opposed to the 13,549 derived from manual annotations. 
MegaDetector needs no changes or special training to do this. 
We see that the performance increases compared to using fewer detections. 
Finally, we tried an alternative version of the above experiment where we added detections from the held out locations in the test set (denoted as MDx2 v2), for the same total of 25,406 cropped images. 
Again we see an additional improvement but without increasing further the number of images, indicating that the downstream task performance can benefit from pretext training with images from similar locations with the test set. 

We conclude that the SSL methods evaluated are robust to how the training cropped images are generated and result in performance that is comparable to using manually annotated boxes. 
An obvious question is what would the performance be like if you only used entire images and not ones cropped around the objects of interest. 
Existing work has shown that cropped images are much more effective in the case of image classification from camera traps~\cite{beery2018recognition}. 
Given this and the availability of highly accurate detectors, we chose not to address this question. 

\subsection{What is the impact of increasing model capacity?}

All experiments in the main paper were conducted using a ResNet18~\cite{he2016deep} as the backbone feature extractor with an image resolution of $112\times112$ pixels. 
As we use images cropped around the objects of interest as input, this resolution is more than sufficient for capturing the visually important characteristics of the categories. 

In Table~\ref{tab:depth_and_resolution} we show the impact of increasing both the backbone model capacity (\ie using a ResNet50) and the image resolution (\ie using an image side of size $224$).
As expected, we see performance improvements across all conditions, but importantly the ranking of the methods is relatively stable. 
We conclude that for best possible performance, not surprisingly one should use large models and higher resolution.  
However, this comes at the expense of increased training times and memory consumption.

\subsection{What is the impact of initializing end-to-end supervised training with weights learned from self-supervision?}

The results in Figure~5 of the main paper were computed with the linear evaluation scheme that is commonly used in SSL \ie by training a linear model using self-supervised features as input. 
Here, we use the SSL models as an initialization and fine-tune all the weights of the backbone network to understand if the improvements reported in the linear evaluation case follow through to the end-to-end one. 
The most realistic setting for camera-trap data is the low label regime where only a small  number of images have been annotated. 
With this in mind, we only present results for the 1\% and 10\% labeled images settings. 
When performing supervised training, all hyper parameters are the same as the ones we use for SSL with the exception of the learning rate which we decrease to 0.003.
In Table~\ref{tab:end_to_end} we observe that in almost all cases, models that are initialized with SSL derived weights are vastly superior to those that use only ImageNet initialization (denoted as Standard). 
Again, we see that the models trained with context result in a much better initialization than without. 

\subsection{What happens if we start SSL from random weight initialization?} 

All the results in the main paper use models that have been pretrained on ImageNet. 
While it is more common in SSL to start from randomly initialized weights, we instead adopted the more pragmatic viewpoint that pretrained networks are readily available, and thus practitioners are likely to use them as a starting point.  
In the interest of making progress on camera trap image classification (where lack of image labels is the main issue), we chose to start from ImageNet pretrained models. 
Not surprisingly, if we initialize our SSL models with random weights, the performance is worse on the medium-sized datasets that we use for our experiments (see Table~\ref{tab:random_init}). 
Importantly, we still observe that utilizing context information is superior to standard augmentation-based SSL. 

\subsection{Are accuracy gains concentrated with the majority-categories, or spread across multiple categories?} 

\begin{figure*}[t]
\begin{center}
 \includegraphics[width=0.9\linewidth]{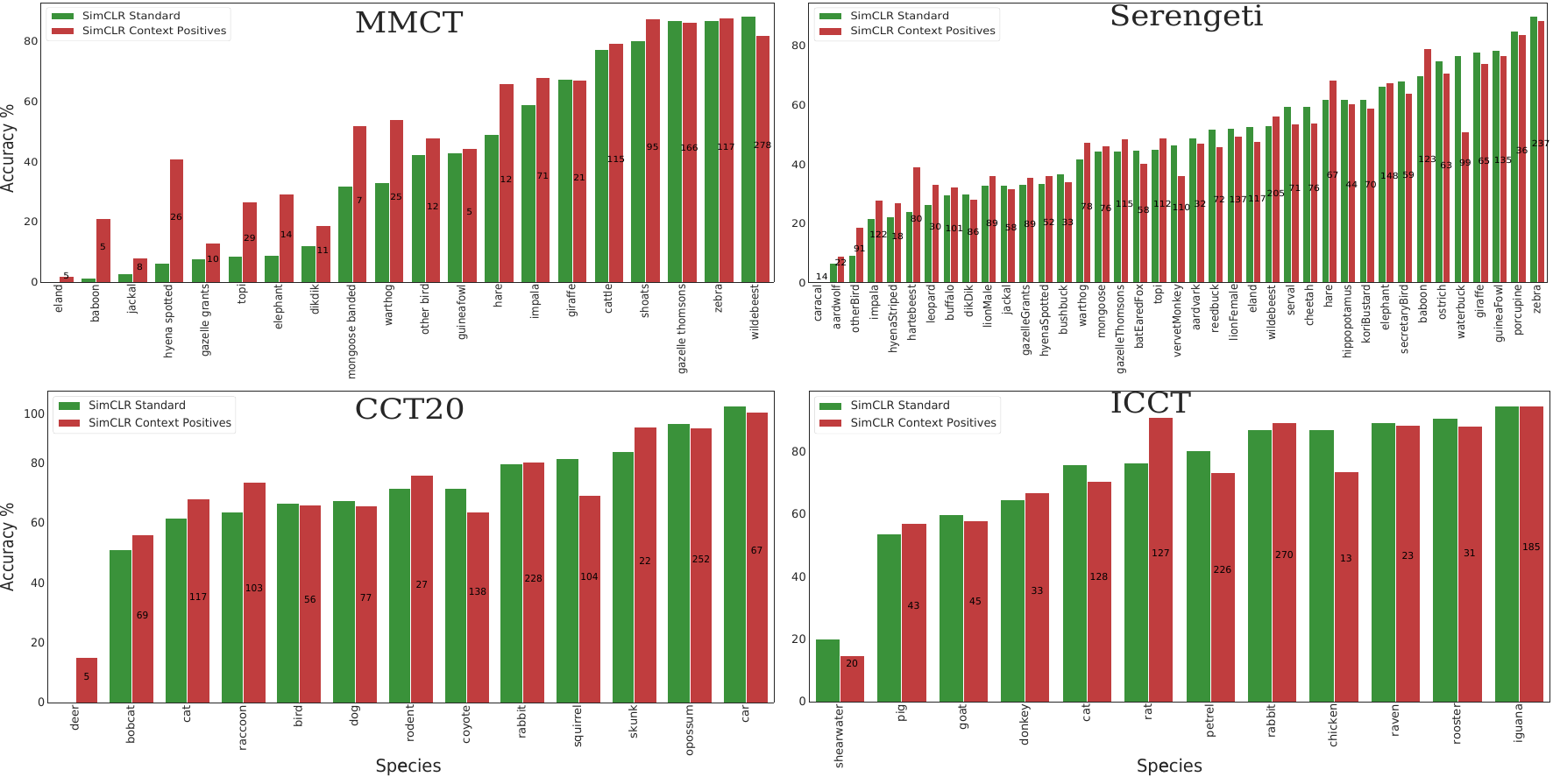}
\end{center}
   \vspace{-5pt}
   \caption{
   Evaluating per-class accuracy. Here we report test-time,  per-class, linear evaluation accuracy for each of our four different datasets. 
   We compare standard image augmentation to our context-based approach using SimCLR as the SSL algorithm. 
   }
   \vspace{-10pt}
\label{fig:species_accuracy}
\end{figure*}

\begin{figure*}[h] 
\begin{center}
 \includegraphics[width=0.9\linewidth]{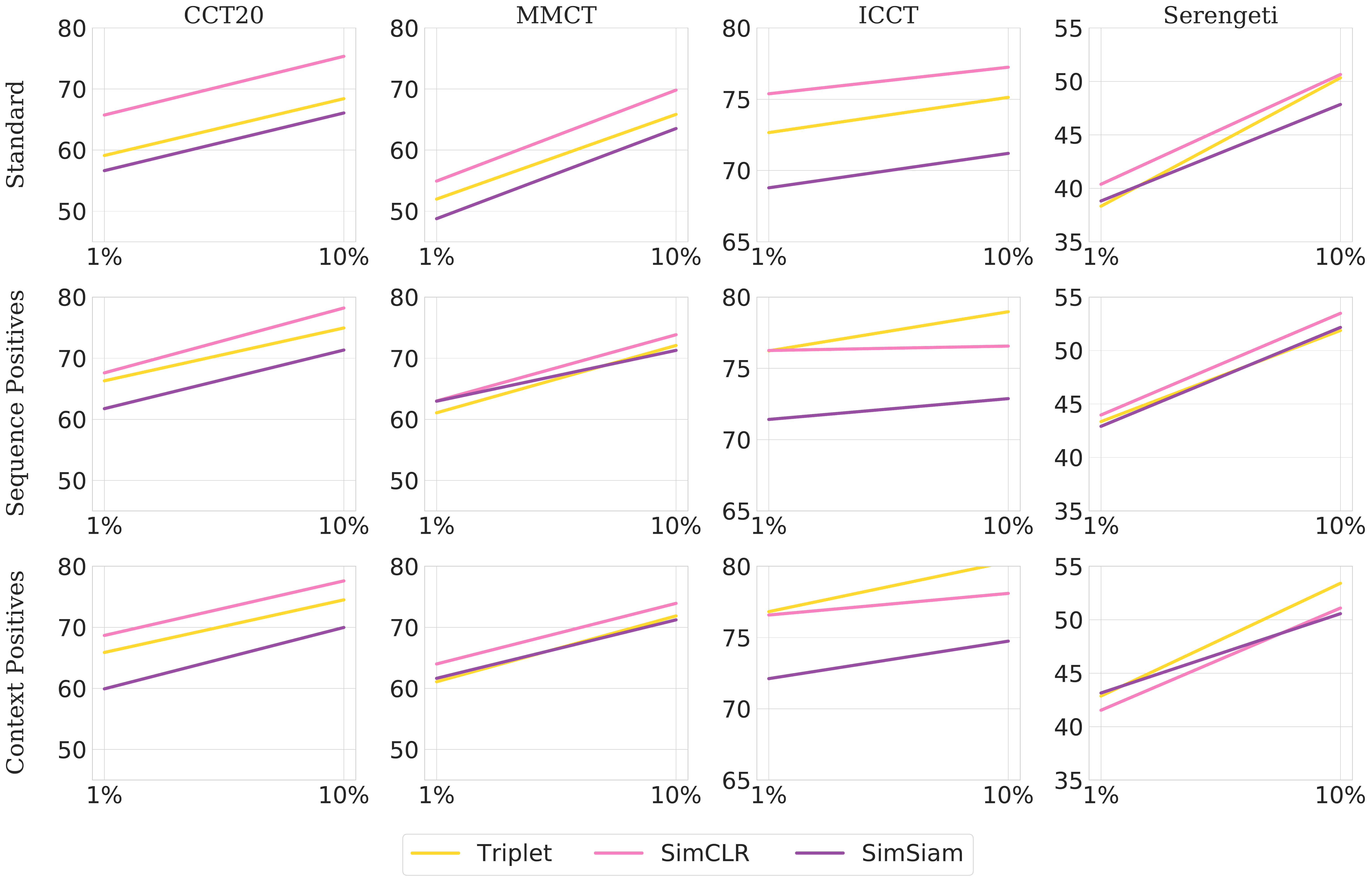}
\end{center}
   \vspace{-10pt}
   \caption{
   Linear evaluation accuracy of different SSL approaches and positive pair selection strategies. 
   This is an alternative presentation of the results from Figure~5 from the main paper where it is easier to compare the different SSL methods on a given dataset.
   }
   \vspace{-5pt}
\label{fig:method_comparison_results}
\end{figure*}

In Figure~\ref{fig:species_accuracy} we compare the per-category, 10\% linear evaluation, accuracy of standard SimCLR and SimCLR with our context-based positive based sampling. 
For each category, we also include the number of examples in the 10\% subset of the training data.
There is no dominant pattern, and we see that context-based sampling helps for both well-represented and under-represented categories. 

\subsection{What is the proposed algorithm getting right, that ``Standard'' SSL is getting wrong?}

\begin{figure*}[h] 
\begin{center}
 \includegraphics[width=1.0\linewidth]{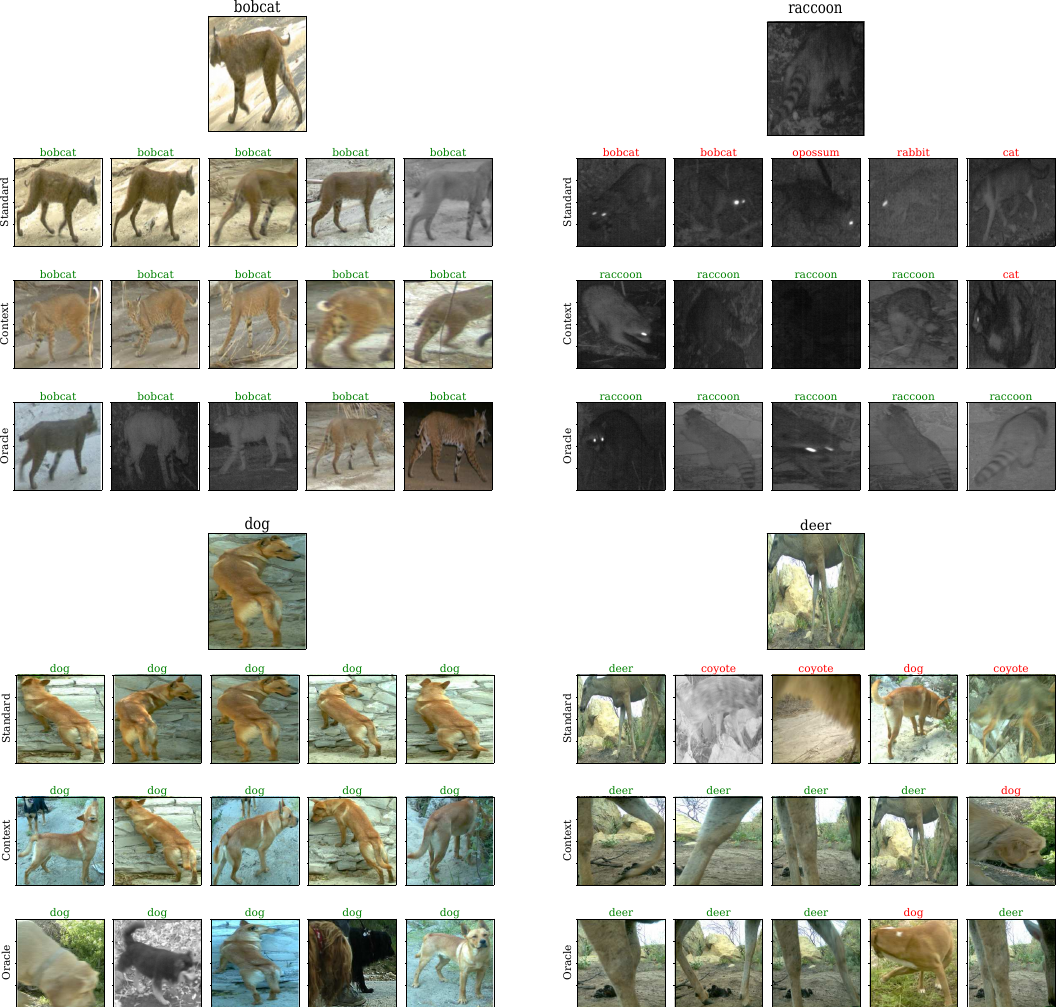}
\end{center}
   \caption{
   Nearest neighbor retrieval results for SimCLR for four different test images from CCT20. 
   For each of the different images, we show the top five nearest neighbors in 128 dimensional embedding space (\ie the output of the projector) for three different SimCLR models.
   The label on top of each image is the ground truth class name.
   We see that the nearest neighbors for `Standard' SimCLR display very limited visual diversity.
   The unobtainable `Oracle' model, which has been trained with ground truth labels, has the most variety. 
   Our `Context' approach is between the two extremes and shows non-trivial diversity which indicates that it contains more semantic information in it's features compared to conventional augmentation-based SimCLR. 
   Note, that none of these images have been observed during self-supervised training.
   }
   \vspace{-5pt}
\label{fig:qualitative_results}
\end{figure*}

\begin{figure*}[h] 
\begin{center}
 \includegraphics[width=1.0\linewidth]{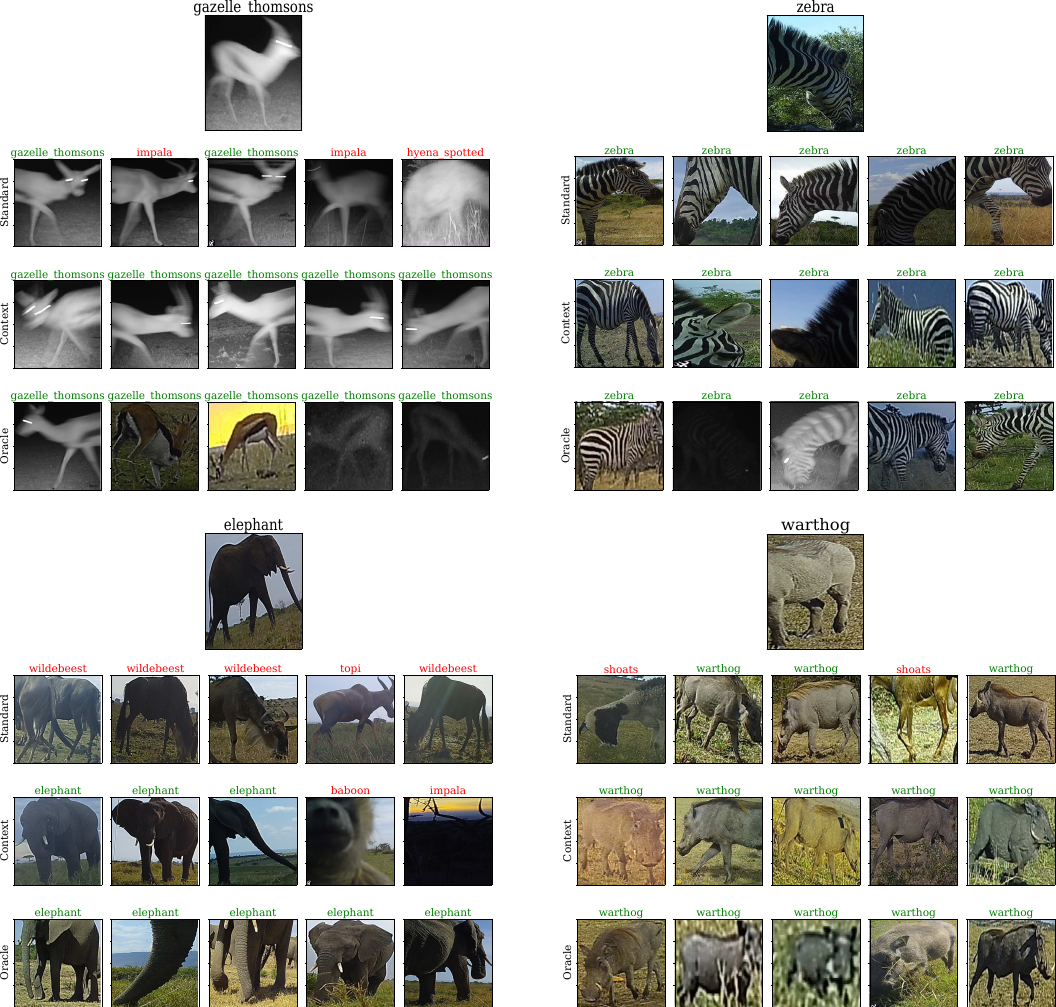}
\end{center}
   \caption{
   Nearest neighbor retrieval results for SimCLR for four different test images from MMCT. 
   For each of the different images, we show the top five nearest neighbors in 128 dimensional embedding space (\ie the output of the projector) for three different SimCLR models.
   The label on top of each image is the ground truth class name. Again, our ‘Context’ approach captures more diverse appearances and perspectives of animals compared to conventional augmentation-based SimCLR. Note, that none of these images have been observed during self-supervised training.
   }
   \vspace{-5pt}
\label{fig:qualitative_results_kenya}
\end{figure*}

In order to attempt to address this question, we implemented a simple nearest neighbor retrieval visualization. 
A given query image is used to retrieve the five nearest neighbors in embedding space. 
In Figure~\ref{fig:qualitative_results} we show example results on CCT20. 
We can see that the standard augmentation-based SimCLR model leads to retrieved images that look qualitatively like the query in that they have similar lighting and orientation (\eg portrait \vs landscape). 
But they sometimes contain the wrong animal species, compared to ours (with `Context'), which seems to capture more diverse appearances and better emphasize species-related characteristics. We observe similar trends for MMCT in Figure~\ref{fig:qualitative_results_kenya}. 

The top-left example in Figure~\ref{fig:qualitative_results} and the top two examples in Figure~\ref{fig:qualitative_results_kenya} illustrate a limitation of both of the self-supervised methods. 
We can see that the oracle is capable of retrieving images with large illumination changes (\ie spanning night and day). 
In these examples, it appears that the self-supervised methods are not able to merge these distinct visual modes. 
This is despite the large amount of color augmenting that they are exposed to during training \eg color jittering and grayscale conversion.   
An interesting future question, is what additional information can we make use of during training to merge these diverse modes within a given category. 

\subsection{Is the proposed approach applicable beyond camera trap data?}

While we believe that the four quite distinct camera trap datasets explored in the paper constitute an important problem that deserves dedicated attention, we explore the applicability of our approach on the Functional Map of the World (FMoW) \cite{christie2018functional} to test the generalizability of the findings.  FMoW is a satellite imagery dataset that contains annotated images of categories relevant to the functional purpose of buildings or land use. The data comes in temporal sequences and is accompanied by metadata which make them suitable for validating our approach. We used a subset of the data that consists of 30 different classes, with 30,014 images reserved for training and 10,085 for testing.
The results in Table~\ref{tab:results_fmow}, show that: (i) SSL is superior to ImageNet features and (ii) our context selection is consistently better than standard SSL, especially in the low-data regime.

\section{Implementation Details} 
Here we provide additional implementation details related to our experimental evaluation. 

\subsection{Training} 

Unless otherwise stated, each SSL network uses a backbone initialized with ImageNet weights. 
We train all models for 200 epochs with a batch size of 256 and a learning rate of 0.03, using a cosine annealing schedule. 
We use SGD with momentum of 0.9 and weight decay of 0.0005. 
The projector $g$ is a two layer MLP with a hidden layer  of size 512 and size 128 for the output. 
The predictor $h$, used by SimSiam, is also a two layer MLP with a hidden layer of size 64. 
For SimSiam only, as in the original paper~\cite{chen2021exploring}, we add batch normalization to the output of the first layer for the projector and predictor -- the model performed poorly without it. 
For the larger capacity models, we reduced the batch size by half and scaled the learning rate by half also.

For the triplet loss in Equation 1 we set the margin to 0.3.
To scale the distances used by SimCLR, we use a temperature of 0.5, see Equation 2 in the main paper.  
The context temperature parameter in Equation 5 is set to 0.05. 

For the sequence positive approach in Section 3.2, we consider all images that are from the same location that are captured within 5 seconds of each other as potential positives.

We use the same set of augmentations for all SSL and end-to-end supervised methods at training time. 
This includes random resized crops in the range $[0.2, 1.0]$, horizontal flipping with probability $0.5$, color jittering with probability $0.8$, and grayscale conversion with probability $0.2$. 

\subsection{Evaluation} 
When training the linear classifiers for evaluation we use logistic regression with L2 regularization as implemented in \texttt{scikit-learn} \footnote{scikit-learn: \url{https://scikit-learn.org}}.
We use a lbfgs solver, with a maximum number of iterations of 1000 and a multinomial loss. 
To select the regularization weighting, we search over the set \{$0.001, 0.01, 0.1, 1, 10, 100$\}, and choose the best value using five-fold cross validation on the training set.
When training the linear classifier, we simply resize the cropped image to the desired image resolution (\eg $112\times 112$) when extracting features, and we do not use any other augmentation. 

When generating the 1\% and 10\% subsets, we randomly sampled the corresponding percentage of images of each class in the full training set, while also ensuring that there were at least one example per class.
Some categories are more common than others, so this sub-sampling procedure preserves the imbalance that is typical in camera trap datasets. 
We use the same fixed subsets for all experiments for a given dataset. 

\begin{table}[h]
\centering
\footnotesize
\begin{tabular}{ |c|c|c|c|c| }
\hline
\multicolumn{5}{|c|}{\bf Functional Map of the World (FMoW)}\\
\hline
Approach & Method  & 1\% & 10\% & 100\% \\ 
\hline
Supervised End-to-End\text{*}
    &- &37.62 &51.79 &58.16\\
\hline
Supervised
    &-  &36.23 &46.24 &52.72\\
 \hline
Triplet
    & Standard  &44.38 &54.05 &54.58 \\ 
    & Seq. Pos. &\textbf{50.66} &\textbf{55.22} &56.67 \\  
    & Con. Pos. &49.75 &55.16 &\textbf{57.28}\\ 
 \hline
SimCLR 
    & Standard &45.60 &51.53 &52.16 \\  
    & Seq. Pos. &45.30 &50.07 &50.85 \\  
    & Con. Pos. &\textbf{45.93} &\textbf{52.24} &\textbf{53.03}\\ 
\hline
SimSiam
    & Standard  &46.58 &51.74 &\textbf{55.76} \\  
    & Seq. Pos. &46.63 &51.50 &52.97 \\ 
    & Con. Pos. &\textbf{47.64} &\textbf{53.51} &52.80\\
    \hline
\end{tabular}
\vspace{+5pt}
\caption{Linear evaluation of our approach on a subset of FMoW, a satellite imagery dataset. Comparison of SSL approaches shows that using context benefits in most cases, especially in the low-data regime. The results here are averaged over three runs. \text{*}We also include fully supervised end-to-end training for reference.}
\vspace{+5pt}
\label{tab:results_fmow}
\end{table}

\begin{table*}[h]
\centering
\small
\begin{tabular}{ |c|c|cccc|cccc| }
\hline
\multicolumn{10}{|c|}{\bf Maasai Mara Camera Traps (MMCT)}\\
\hline
 & & 
 \multicolumn{4}{c|}{1\%} &
 \multicolumn{4}{c|}{10\%}  \\ \cline{3-10}
Approach & Method  & GT & MD & MDx2 v1 & MDx2 v2 & GT & MD & MDx2 v1 & MDx2 v2\\ 
\hline
Triplet
    & Standard &51.99 &52.93 &55.35 &55.76 &65.86 &64.85 &68.47 &68.43\\ 
    & Seq. Pos. &61.06 &60.63  &\textbf{61.63} &62.82 &\textbf{72.09} &70.79 &72.27 &73.25\\  
    & Con. Pos. &\textbf{61.10} &\textbf{60.79} &61.51 &\textbf{63.52} &71.88
 &\textbf{70.88} &\textbf{72.66} &\textbf{73.27} \\
 \hline
SimCLR 
    & Standard &54.93 & 56.30 &59.46 &60.32 &69.83 &68.95 &72.67 & 73.46\\ 
    & Seq. Pos. &63.00 &62.55 &\textbf{64.11} &64.57 &73.84 &72.21 &\textbf{74.37} &74.37\\  
    & Con. Pos. &\textbf{64.01} &\textbf{63.84} &63.98 &\textbf{67.29} &\textbf{73.93} &\textbf{72.77} &74.03 &\textbf{76.01}\\ 
\hline
SimSiam
    & Standard &48.78 &47.87 &56.75 &57.17 &63.53 &62.98 &68.33 &68.11\\
    & Seq. Pos. &\textbf{62.96} &61.18 &63.08 &\textbf{66.90} &\textbf{71.28} &70.46 &72.29 &\textbf{74.73} \\
    & Con. Pos. &61.66 &\textbf{62.78} &\textbf{64.83} &66.59 &71.23 &\textbf{70.78} &\textbf{72.40} &74.01 \\ 
    \hline
\end{tabular}
\vspace{+5pt}
\caption{Comparing detected bounding boxes from MegaDetector (MD) with ground truth (GT) boxes on the MMCT dataset. 
We also evaluate the effect of doubling the camera trap detections from locations that are not in the test set (MDx2 v1) and locations that belong to the test set (MDx2 v2). }
\vspace{+5pt}
\label{tab:mega_detector}
\end{table*}

\begin{table*}[h]
\centering
\small
\begin{tabular}{ |c|c|cc|cc|cc| }
\hline
\multicolumn{8}{|c|}{\bf Caltech Camera Traps (CCT20)}\\
\hline
 & & 
 \multicolumn{2}{c|}{1\%} &
 \multicolumn{2}{c|}{10\%} &
 \multicolumn{2}{c|}{100\%} \\ \cline{3-8}

Approach & Method  & RN18-112 & RN50-224 & RN18-112 & RN50-224 & RN18-112 & RN50-224\\ 
\hline
Triplet
    & Standard  &59.13 &61.99 &68.43 &73.85 &73.40 &80.10\\ 
    & Seq. Pos. &\textbf{66.30} &70.80 &\textbf{74.96} &80.96 &\textbf{77.31} &\textbf{83.67}\\  
    & Con. Pos. &65.90 &\textbf{70.87} &74.51 &\textbf{81.22} &76.42 &83.11\\
 \hline
SimCLR 
    & Standard  &65.75 &68.59 &75.36 &78.98 &76.37 &83.39\\ 
    & Seq. Pos. &67.60 &74.96 &\textbf{78.22} &82.74 &\textbf{78.99} &\textbf{85.54}\\  
    & Con. Pos. &\textbf{68.67} &\textbf{75.06} &77.60 &\textbf{82.78} &78.02 &85.14\\ 
\hline
SimSiam
    & Standard  & 56.64 &57.47 &66.08 &72.42 &72.13 &78.61\\
    & Seq. Pos. & \textbf{61.74} &66.16 &\textbf{71.34} &78.39 &\textbf{74.78} &81.52\\
    & Con. Pos. & 59.93 &\textbf{69.95} &69.98 &\textbf{79.14} &74.38 &\textbf{82.59}\\ 
\hline
\end{tabular}
\vspace{+5pt}
\caption{Evaluating model capacity. Linear evaluation accuracy on CCT20 with increased model capacity from ResNet18 (RN18) to ResNet50 (RN50) and image resolution from $(112\times112)$ to $(224\times224)$. We compare the positive-pair mining methods across all SSL approaches.}
\vspace{+5pt}
\label{tab:depth_and_resolution}
\end{table*}

\begin{table*}[h]
\centering
\small
\begin{tabular}{ |c|c|c|cc|cc| }
\hline
 & & &
 \multicolumn{2}{c|}{1\%} &
 \multicolumn{2}{c|}{10\%}\\ \cline{4-7}
Dataset & Approach & Method  & Lin. Eval. & End-to-End & Lin. Eval & End-to-End \\ 
\hline
CCT20
 &Supervised &-  &- &55.65 &- &75.14\\  
 &SimCLR  &Standard  &65.75 &68.76 &75.36 &78.20\\ 
 &SimCLR  &Seq. Pos. &67.60 &\textbf{72.56} &\textbf{78.21} &\textbf{79.33}\\  
 &SimCLR  &Con. Pos. &\textbf{68.67} &72.06 &77.61 &78.73 \\
 \hline
MMCT
    &Supervised &-  &- &53.00 &- &67.73\\ 
    &SimCLR &Standard  &54.93 &59.30 &69.82 &73.97\\ 
    &SimCLR &Seq. Pos. &63.00 &\textbf{62.91} &73.84 &74.22\\  
    &SimCLR &Con. Pos. &\textbf{64.00} &62.23 &\textbf{73.93} &\textbf{75.89}\\ 
\hline
ICCT
    &Supervised &-  &- &62.50 &- &76.60 \\ 
    &SimCLR &Standard  &75.38 &75.14 &77.24 &\textbf{79.44}\\
    &SimCLR &Seq. Pos. &76.26 &76.02 &76.56 &77.38\\
    &SimCLR &Con. Pos. &\textbf{76.58} &\textbf{77.13} &\textbf{78.10} &79.12\\ 
\hline
Serengeti
    &Supervised &-  &- &36.03 &- &\textbf{55.29}\\ 
    &SimCLR &Standard  & 40.37 &41.30 &50.64 &54.17\\
    &SimCLR &Seq. Pos. &\textbf{43.97} &\textbf{43.79} &\textbf{53.48} &54.58\\
    &SimCLR &Con. Pos. &41.53 &42.62 &51.10 &54.19
\\ 
\hline
\end{tabular}
\vspace{+5pt}
\caption{Comparing linear versus end-to-end supervised finetuning. Starting from SimCLR derived self-supervised representations, we compare linear evaluation (as in the main paper) to end-to-end supervised finetuning from SSL initialization. 
`Supervised' refers to the performance of the fully-supervised baseline, initialized from ImageNet only, without using SSL. }
\vspace{+5pt}
\label{tab:end_to_end}
\end{table*}

\begin{table*}[h]
\centering
\small
\begin{tabular}{ |c|c|cc|cc|cc| }
\hline
\multicolumn{8}{|c|}{\bf Caltech Camera Traps (CCT20)}\\
\hline
 & & 
 \multicolumn{2}{c|}{1\%} &
 \multicolumn{2}{c|}{10\%} &
 \multicolumn{2}{c|}{100\%} \\ \cline{3-8}
Approach & Method  & ImageNet & Random & ImageNet & Random & ImageNet & Random\\ 
\hline
Triplet
    & Standard  &59.13 &45.22 &68.43 &51.04 &73.40 &56.14\\ 
    & Seq. Pos. &\textbf{66.30} &46.53 &\textbf{74.96} &\textbf{55.24} &\textbf{77.31} &\textbf{59.35}\\  
    & Con. Pos. &65.90 &\textbf{46.90}  &74.51 &54.65 &76.42 &59.19\\
 \hline
SimCLR 
    & Standard  &65.75 &53.29 &75.36 &62.90 &76.37 &66.89\\ 
    & Seq. Pos. &67.60 &58.09 &\textbf{78.22} &\textbf{67.40} &\textbf{78.99} &\textbf{70.84}\\  
    & Con. Pos. &\textbf{68.67} &\textbf{58.86} &77.60 &67.36 &78.02 &69.41\\ 
\hline
SimSiam
    & Standard  &56.64 &41.36 &66.08 &48.77 &72.13 &53.45\\
    & Seq. Pos. &\textbf{61.74} &43.04 &\textbf{71.34} &50.70 &\textbf{74.78} &54.92\\
    & Con. Pos. &59.93 &\textbf{43.82} &69.98 &\textbf{51.35} &74.38 &\textbf{56.27}\\ 
    \hline
\end{tabular}
\vspace{+5pt}
\caption{Comparing random initialization to ImageNet initialization. Linear evaluation comparison of SSL approaches on CCT20 with the backbone feature extractor $f$  initialized with weights either from ImageNet (as in the main paper) or randomly.}
\vspace{+5pt}
\label{tab:random_init}
\end{table*}

\end{document}